\documentclass[runningheads]{llncs}

\usepackage{eccv}

\usepackage{eccvabbrv}

\usepackage{graphicx}
\usepackage{booktabs}
\usepackage{enumitem}
\usepackage{wrapfig}

\usepackage[accsupp]{axessibility}  

\usepackage{hyperref}

\usepackage{orcidlink}
\definecolor{flowblue}{RGB}{102, 203, 250}
\definecolor{flowred}{RGB}{235, 73, 52}

\begin{document}

\title{milliFlow: Scene Flow Estimation on mmWave Radar Point Cloud for
Human Motion Sensing} 

\titlerunning{milliFlow}

\author{Fangqiang Ding\inst{1}\orcidlink{0000-0001-5287-7128} \and
Zhen Luo\inst{1}\orcidlink{0009-0006-7815-7464} \and
Peijun Zhao\inst{2}\orcidlink{0000-0003-2843-6941
} \and
Chris Xiaoxuan Lu\inst{3}\thanks{Corresponding author. Email: xiaoxuan.lu@ucl.ac.uk}\orcidlink{0000-0002-3733-4480}
}

\authorrunning{F. Ding et al.}

\institute{\textsuperscript{1}University of Edinburgh\hspace{2mm} \textsuperscript{2}MIT\hspace{2mm} \textsuperscript{3}UCL
}

\maketitle

\begin{abstract} 
Human motion sensing plays a crucial role in smart systems for decision-making, user interaction, and personalized services. Extensive research that has been conducted is predominantly based on cameras, whose intrusive nature limits their use in smart home applications. To address this, mmWave radars have gained popularity due to their privacy-friendly features.  In this work, we propose milliFlow, a novel deep learning approach to estimate scene flow as complementary motion information for mmWave point cloud, serving as an intermediate level of features and directly benefiting downstream human motion sensing tasks. Experimental results demonstrate the superior performance of our method when compared with the competing approaches. Furthermore, by incorporating scene flow information, we achieve remarkable improvements in human activity recognition and human parsing and support human body part tracking. Code and dataset are available at \url{https://github.com/Toytiny/milliFlow}. 
  
\end{abstract}

\section{Introduction}

Perceiving and understanding human behaviours play a pivotal role in human-centred applications such as disaster response~\cite{tariq2018dronaid,mishra2020drone}, surveillance~\cite{paul2013human,baltieri20113dpes} and health monitoring~\cite{mukhopadhyay2014wearable,del2015tracking,fagert2019gait}. 
Conventional methods rely on cameras~\cite{guler2018densepose,shu2020host} or wearables~\cite{lemmens2015recognizing,wang2018rf,bannis2023idiot}, which are prone to visual deterioration (such as low lighting conditions, smoke, and fog) and raise privacy concerns, potentially compromising user experience with measurements that are psychologically intrusive. To address these concerns, researchers, on the other side, also propose to use wireless radio frequency (RF) signals bounced off the human body for human sensing \cite{tian2018rf,kianoush2016device,zhang2016wifi,zhang2020data,zhang2019wienhance} which is robust against poor lighting, privacy-preserving and non-intrusive to users. Among many RF techniques, single-chip millimetre wave (mmWave) radar emerges as a low-cost sensor that can provide more trustworthy point clouds of a scene under environment dynamics due to its MIMO transceiver design. For these reasons, there has been a significant increase in the production of single-chip radars~\cite{iwr6843} and wide deployment in real-world scenarios, ranging from smart buildings~\cite{wholehome, tseng2014application} to vehicle cabins \cite{schwarz2021heartbeat,chen2021cabin}, and to first-responder toolkits~\cite{rescueproject}.

\begin{figure}[tbp!]
    \centering
    \includegraphics[width=0.618\textwidth]{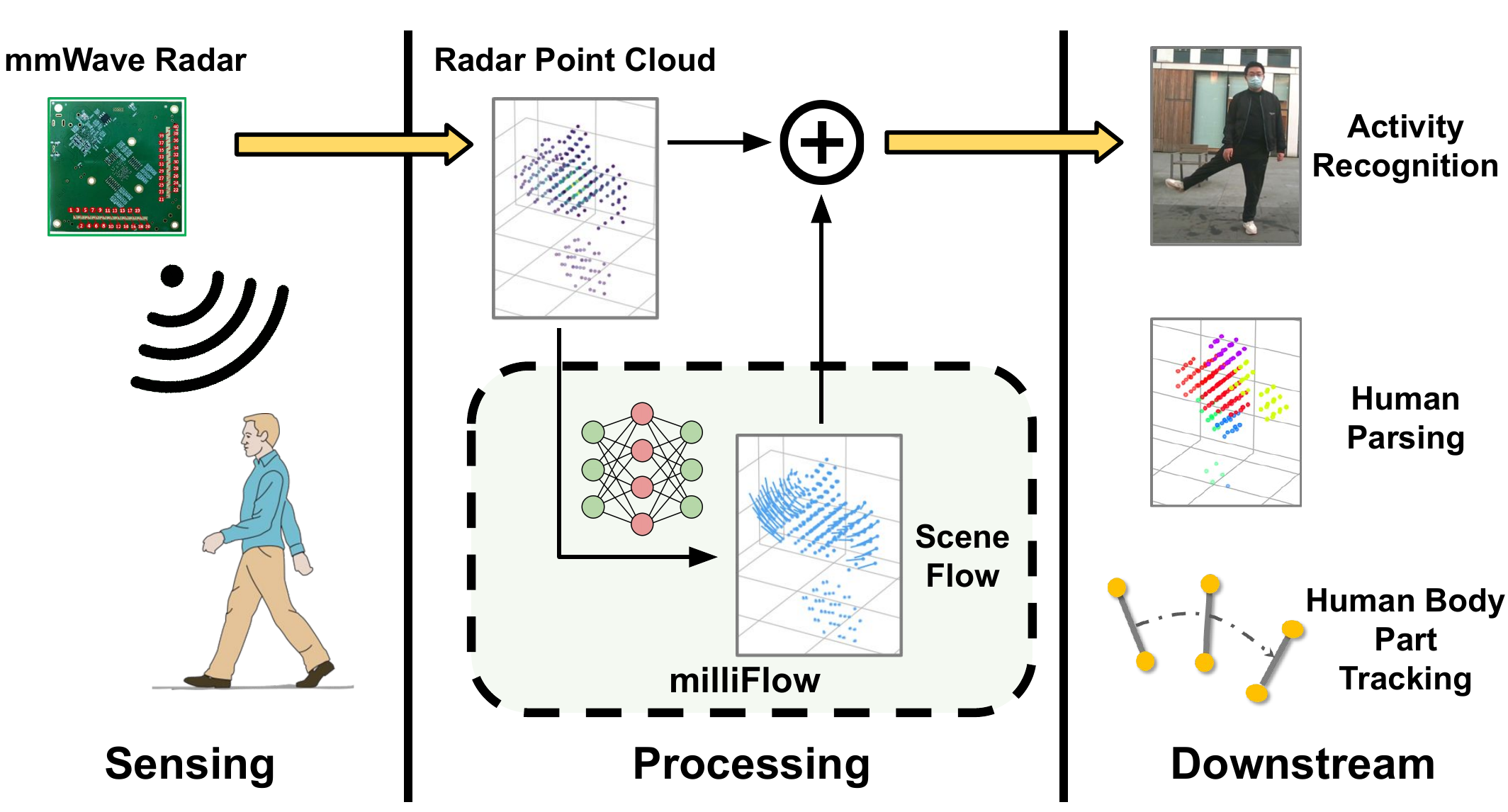}
    \caption{\small{We propose milliFlow, a scene flow estimation module to provide an additional layer of point-wise motion information on top of the original mmWave radar point cloud in the conventional mmWave-based human motion sensing pipeline.}}
    \label{fig:system}
\end{figure}

As motion is naturally continuous and human bodies are non-rigid, point-wise velocity per radar frame is intuitively a strong cue for improving the motion estimation robustness and has been widely used in prior arts~\cite{xue2021mmmesh,kong2022m3track,wang2023human,singh2019radhar}. 
However, fine-grained velocity is difficult to obtain when it comes to mmWave radars. First, while some radars provide the Doppler velocity of a point, the velocity resolution is rather low~\cite{iovescu2017fundamentals} and thus cannot accurately capture the subtle human body movement, which is usually slower than 0.5m/s in domestic life.
Moreover, mmWave radar only senses radial Doppler information but fails to capture the tangential one. For some radars designed for human sensing, e.g., the Vayyar radar~\cite{vayyarhome} used in this work, the Doppler information is even absent in their sensor outputs. Second, extracting motion information from consecutive mmWave radar point clouds is also highly error-prone due to the limitations inherent in low-cost single-chip mmWave radar. These limitations include extremely sparse point clouds due to the target detector, e.g., constant false alarm rate (CFAR) algorithm~\cite{scharf1991statistical} used on-chip and the presence of ghost points caused by multi-path effect~\cite{gennarelli2014multipath}. More recently, it has been found that in a single radar frame, only a subset of body parts reflecting the signal towards the radar can be observed, while other parts deflecting the signal away from the radar are missing from the capture~\cite{cao2022cross}. As a result, some human body parts detected in one frame may disappear in the next frame. For the above reasons, conventional radar point tracking methods~\cite{rohal2017radar,yan2022radar,godrich2010target} struggle to track human motion across frames or give fine-grained velocity in between. 

In this work, we propose to estimate and use scene flow as intermediate features to better support radar-based human sensing. Scene flow refers to a set of displacement vectors between two consecutive point-cloud frames describing the motion field of a 3D scene. We hypothesise that scene flow, if estimated accurately, is able to drastically facilitate cross-frame movement analysis by directly exposing per-point motion features, thereby addressing the aforementioned challenges in previous research. Scene flow has long been proven very effective by the computer vision community in image-based human motion sensing applications~\cite{wang2017scene, liu20123d,jaimez2015motion}. However, estimating scene flow on mmWave radar point cloud is non-trivial because of the inherent sparsity and noise of radar point cloud. Directly applying conventional scene flow estimation methods designed for LiDAR or RGB-D cameras~\cite{wei2021pv,wu2020pointpwc,cheng2022bi,kittenplon2021flowstep3d} on radar point clouds has been found inadequate and generalises poorly. On the other side, recent mmWave radar scene flow estimation works~\cite{ding2022raflow,ding2023hidden} generally focus on autonomous driving scenarios and rely on deep learning-based pipelines. However, these existing methods cannot be readily applied to our human sensing scenarios because the rigid-body assumption used for autonomous driving scenarios cannot stand in our cases where human subjects have non-rigid motion. 

To cope with the above problems, we propose \emph{milliFlow}, as exhibited in \cref{fig:system}, a novel mmWave radar-based scene flow estimation approach for human motion sensing scenarios. Our contributions include:
\begin{itemize}[leftmargin=*,label=$\bullet$]
\setlength{\itemsep}{0pt}
\setlength{\parsep}{0pt}
\setlength{\parskip}{0pt}
\item To the best of our knowledge, milliFlow is the first-of-its-kind work that aims to estimate mmWave radar-based scene flow for human motion sensing.
\item We address the challenges, e.g., sparsity, and lack of temporal cues, for scene flow learning in our cases with a bespoken end-to-end learning network.
\item We propose a cross-modal automatic scene flow labelling scheme specific for human motion sensing, avoiding labour-intensive manual labelling. 
\item We collect a large-scale human motion sensing dataset for evaluation, and perform a comprehensive evaluation of scene flow estimation accuracy as well as the performance on three downstream tasks.
\end{itemize}

\section{Related Work}

\noindent\textbf{mmWave Radar-based Human Sensing.}
The feasibility and versatility of mmWave radar have been extensively demonstrated in various human sensing applications, including vital sign monitoring~\cite{alizadeh2019remote,wang2020remote,lv2021non}, signature verification~\cite{han2023mmsign,li2023mmhsv}, fall detection~\cite{jin2020mmfall,wang2020millimetre}, human tracking and identification~\cite{zhao2019mid,zhao2021human,cui2021high, 10.1145/3610902,gu2019mmsense}, gesture recognition~\cite{liu2021m,hazra2018robust,10.1145/3448110,10.1145/3432235}, activity recognition~\cite{singh2019radhar,wang2021m,ahuja2021vid2doppler} and pose estimation, reconstruction~\cite{xue2021mmmesh,xue2022m4esh,kong2022m3track}. Compared with these applications, our work is unique in that we aim to estimate point-level scene flow vectors instead of providing a holistic output for the whole point cloud. In this way, we can either explicitly augment each radar target with scene flow vectors or implicitly learn robust latent features, which can further benefit many downstream tasks (\cf Sec.~\ref{downstream_eval}). 

\noindent\textbf{Scene Flow Estimation on Point Clouds.} 
Recent scene flow works are mostly towards autonomous driving applications and attempt to estimate scene flow on LiDAR point clouds captured from autonomous vehicles. To this goal, different approaches are proposed, including classical methods~\cite{dewan2016rigid,pontes2020scene,li2021neural,chodosh2024re} and deep learning-based ones~\cite{liu2019flownet3d,baur2021slim,wu2020pointpwc,wei2021pv}. Recently, with the advances in point cloud feature learning, deep learning-based methods become more prevalent.
According to their learning paradigm, prior works in this thread can be divided into fully-supervised~\cite {pan2023moving, wang2020flownet3d++,hermes2023point,wang2021festa,puy2020flot,wei2021pv,behl2019pointflownet}, self-supervised~\cite{kittenplon2021flowstep3d,li2022rigidflow,baur2021slim, mittal2020just,wu2020pointpwc} and weakly-supervised~\cite{gojcic2021weakly,dong2022exploiting} learning approaches. 
Apart from the aforementioned works, our work aims to estimate scene flow for human sensing using mmWave radar. Moreover, our method does not demand any manual annotation efforts, instead, we automatically generate noisy pseudo scene flow labels from corresponding RGB-D images captured by the co-located camera.

\noindent\textbf{Scene Flow Estimation with mmWave Radar.} As far as we know, there are limited works that estimate scene flow using mmWave radar.
A pioneering work~\cite{ding2022raflow} introduced a self-supervised learning method tailored for automotive radar, utilizing unique loss functions and a two-stage network. To improve scene flow performance and enable more downstream applications, a later work~\cite{ding2023hidden} exploits cross-modal supervision from co-located sensors (e.g. IMU, LiDAR) on modern autonomous vehicles for radar scene flow learning. However, these methods cannot be transferred to human sensing scenarios due to two reasons. First, the radar used for human-centric applications is different from the automotive radar in many aspects. For example, automotive radar has a lower range but higher angular resolution, making scene flow models bespoken for them hard to generalize to human sensing. More importantly, the human object is a non-rigid body, while in autonomous driving, scene dynamics are usually attributed to rigid body motion (e.g. cars and motorcycles)~\cite{gojcic2021weakly,dewan2016rigid,dong2022exploiting}. Such discrepancy in objective indicates different label and supervision generation schemes. 

\section{Methodology}

\subsection{Overview} 

We formally formulate our scene flow estimation problem in Sec.~\ref{formulation}. Then we elaborate on the technical challenges of using mmWave radar for scene flow estimation under human motion sensing scenarios in~\cref{chanllenges}.  Sec.~\ref{network} details the design of our neural network, which comprises five sequential modules, tackling the sparsity and noise challenges and compensating for the lack of temporal cues. 
In Sec.~\ref{labelling}, we propose a {cross-modal automatic scene flow labelling} scheme, to efficiently label scene flow for point clouds captured in human sensing scenarios. \cref{loss} further introduces our loss function used for training scene flow network.


\subsection{Problem Formulation}\label{formulation}
Here, we consider the problem of scene flow estimation for dynamic 3D point clouds collected by an mmWave radar sensor used in human sensing scenarios. As a general problem setting, the input to point cloud-based scene flow estimation is two consecutive 3D point clouds $\mathcal{P}=\{p_i\}^{N}_{i=1}$ and $\mathcal{Q}=\{q_i\}^{M}_{i=1}$ captured by the same device and the output is a set of 3D vectors $\mathcal{F}=\{f_i\}^{N}_{i=1}$ that align each point ${p}_i$ in $\mathcal{P}$ to its associated position ${p}^{a}_i = {p}_i + f_i $ in the frame of $\mathcal{Q}$. Different from the task that finds real correspondences between two frames, scene flow estimation only derives per-point 3D displacement for $\mathcal{P}$ and the associate position ${p}^{a}_i$ does \emph{not} essentially overlap with any points in $\mathcal{Q}$. Besides 3D coordinates information, each point may also have additional properties given mmWave radar point clouds as input, such as Doppler velocity or intensity value. Without loss of generality, here we concatenate the per-point incident properties and 3D point coordinates into the 2D matrix and use $\mathcal{X}=\{x_i\}^{N}_{i=1}$, $\mathcal{Y}=\{y_i\}^{M}_{i=1}$ to denote the data from the source and target frame, respectively. 

\subsection{Technical Challenges}\label{chanllenges}

\noindent \textbf{Sparsity and Noise.} 
Due to bandwidth and hardware limitations, the radar raw data has low resolution in both range and angular dimensions from which only outstanding peaks are selected as valid targets. As a result, the point cloud generated by mmWave radar is quite sparse with an average of only around 100 points (in our case) related to the human body, often missing data for specific body parts. Furthermore, the multi-path effect introduces ghost points, adding noise to the already sparse data. This sparsity and noise significantly challenge mmWave radar's ability for scene flow estimation, as the lack of sufficient local geometric cues and the presence of noisy points make it difficult to extract robust local features, essential for accurately tracking movements within the scene.

\noindent \textbf{Lack of Temporal Cues.} 
The radar Doppler velocity measurement, representing the radial velocity of points, could enhance radar scene flow estimation. However, its resolution is limited by hardware and often too low to accurately capture small-scale movements typical in human sensing scenarios. Additionally, some radars, like the Vayyar radar~\cite{vayyarhome} used here, do not measure Doppler velocity due to their specific signal transmitting design, further complicating accurate motion estimation. This limitation, coupled with the absence of consistent radar point data for certain body parts across frames, presents non-trivial challenges for reliable scene flow estimation as they result in the lack of temporal cues.


\noindent \textbf{Scene Flow Annotation.} 
Learning accurate scene flow estimation with deep networks requires point-level scene flow labels for training. However, manual annotation is prohibitively expensive and lacks real-world correspondence. Recent approaches~\cite{jund2021scalable,huang2022dynamic,ding2023hidden} annotate object bounding boxes to generate pseudo scene flow labels, effective in autonomous driving with rigid bodies like cars. Nevertheless, this method is still labour-intensive and less applicable to human sensing due to the non-rigid nature of human movement, posing a challenge in creating accurate training labels for mmWave radar-based scene flow estimation.

\begin{figure*}[!tbp]
    \centering
    \includegraphics[width=\textwidth]{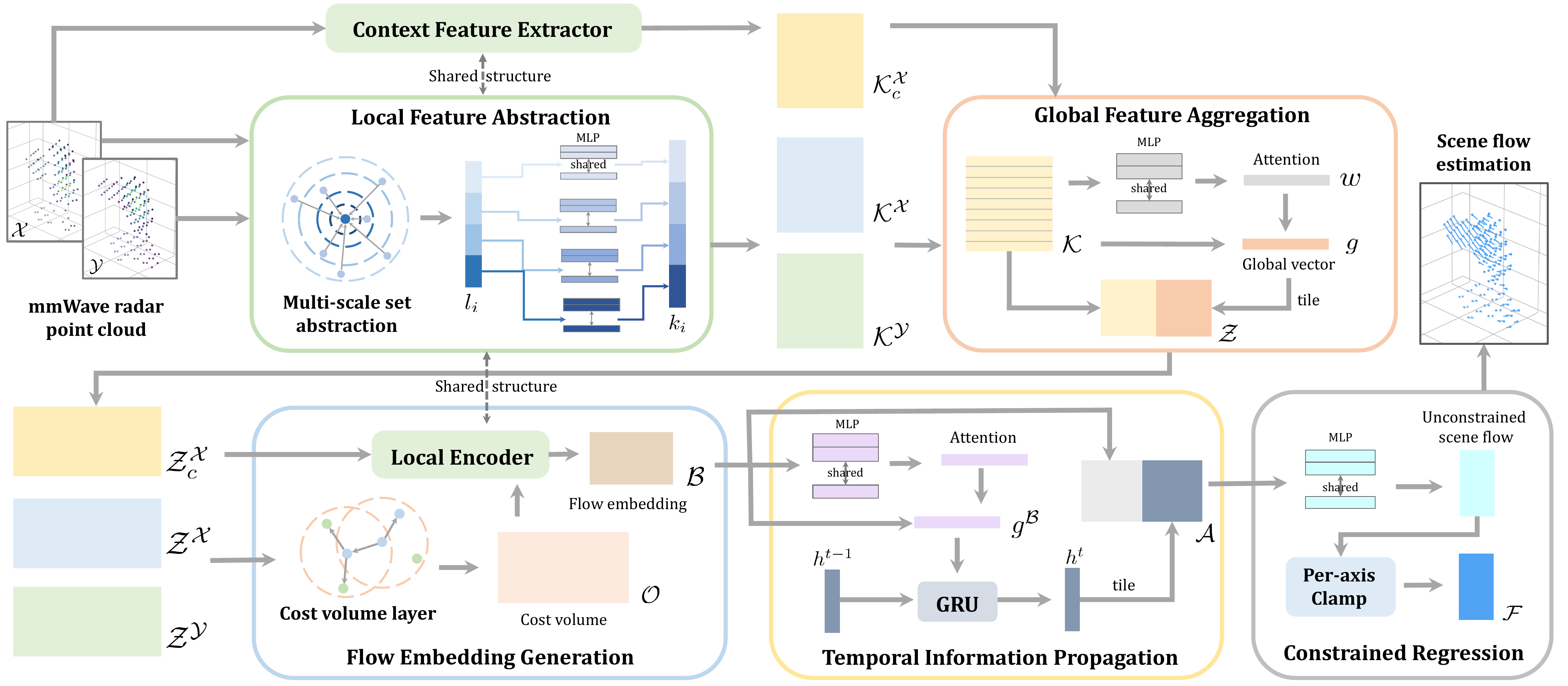}
    \caption{\small{mmWave-based scene flow network architecture. The network takes consecutive radar point clouds as the input and outputs the scene flow in between.  }}
    \label{fig:network}
\end{figure*}

\subsection{Scene Flow Network}
\label{network}

We employ end-to-end trainable deep neural networks for learning point-based scene flow estimation, in line with the state-of-the-art~\cite{ding2023hidden,wei2021pv,kittenplon2021flowstep3d,gojcic2021weakly}. The network architecture is sketched in Fig.~\ref{fig:network}. Particularly, we overcome the sparsity and noise challenges by integrating global features with local ones to provide a comprehensive view of each point cloud. Additionally, we address the lack of Doppler velocity and capture inconsistencies by leveraging a GRU network~\cite{cho2014properties} to incorporate temporal information into the scene flow estimation. In subsequent sections, we will describe the details of each module in the network.


\noindent\textbf{Local Feature Abstraction.}\label{local}
Given mmWave radar point clouds $\mathcal{X}$ and $\mathcal{Y}$ as inputs, we encode their local features using four parallel SA layers~\cite{qi2017pointnet++} with varying grouping radii, enabling multi-scale local feature extraction to account for radar point clouds' non-uniform density. Each SA layer generates a local feature $l_{i,s}$ for a given scale $s$, which is then transformed into a higher-level representation $k_{i,s}$ using a shared-weight MLP with scale-specific parameters. For both point clouds, we concatenate these high-level features across scales to form multi-scale local feature sets $\mathcal{K}^{\mathcal{X}}$ and $\mathcal{K}^{\mathcal{Y}}$ (\cf~\cref{fig:network}). Additionally, a separate context extractor, distinct from but structurally similar to the local encoder, generates context features $\mathcal{K}^{\mathcal{X}}_c$ for $\mathcal{X}$, enhancing the feature representation.


\noindent\textbf{Global Feature Aggregation.}\label{global}
In this module, the local feature vector of each radar point is first transformed into a scalar attention weight $w_i$ using an MLP. These weights are normalized to sum to 1 for numerical stability. The global feature vector $g$ is then derived by a weighted sum of all local features, offering an improvement over max-pooling by dynamically adjusting point-wise weights. This global vector is concatenated with each local feature to form local-global representations $\mathcal{Z}^{\mathcal{X}}, \mathcal{Z}^{\mathcal{Y}}$ for $\mathcal{X}$ and $\mathcal{Y}$. Additionally, a global feature for the context $\mathcal{L}_C^\mathcal{X}$ is obtained through another MLP, resulting in $\mathcal{Z}^{\mathcal{X}}_C$, as seen in~\cref{fig:network}.
\subsubsection{Flow Embedding Generation.}\label{embedding}
Given the local-global features of two radar point clouds, i.e., $\mathcal{Z}^{\mathcal{X}}, \mathcal{Z}^{\mathcal{Y}}$, we use the cost volume layer~\cite{wu2020pointpwc} to compute the correlation between them, as seen in Fig.~\ref{fig:network}. By aggregating the spatial relationship and feature similarities between two frames, the point motions are encoded into the cost volumes, denoted as $\mathcal{O}=\{o_i\}_{i=1}^N$. Thanks to the global feature aggregation, the holistic frame information can also be correlated, which yields more robust and stable costs. To further mix it with the context, we then stack the cost volumes $\mathcal{O}$, context features $\mathcal{L}_C^{\mathcal{X}}$ and pass the features into another local encoder to obtain the flow embedding $\mathcal{B}=\{b_i\}_{i=1}^N$.

\noindent\textbf{Temporal Information Propagation.}\label{temporal}
Directly propagating 2-dimension flow embedding from previous frames suffers from a) high computation overload, and b) the change in the number of points. To overcome these issues, we propose to temporally update its global vector $g^\mathcal{B}$ instead of the flow embedding itself  $\mathcal{B}$ (\cf ~\cref{fig:network}). 
We apply a GRU network~\cite{cho2014properties} to update it as hidden states temporally and obtain the final global representation $h^{t} = \mathrm{GRU}(h^{t-1},g^\mathcal{B}; \theta_g)$, where $h^{t-1}$ is the global representation from the last frame and $\theta_g$ is the GRU network parameters. Lastly, we concatenate this updated global representation to each point in the flow embedding and denote the final features as $\mathcal{A}=\{a_i\}_{i=1}^N$. 

\noindent\textbf{Constrained Scene Flow Regression.}\label{constrained} Given the final features $\mathcal{A}$ produced above, we can use an MLP-based flow regressor to decode per-point scene flow $\mathcal{{F}}=\{{f}_i\}_{i=1}^N$. However, estimating unconstrained scene flow may lead to non-viable results, \eg, the magnitude of the flow vector exceeds the normal scale of human body movement. To constrain our predictions, we propose to clamp the estimated scene flow before returning it. We set a fixed threshold $\epsilon$ and constrain the scene flow on each axis to be within the range of [$-\epsilon, \epsilon$], as shown in~\cref{fig:network}. 
\subsection{Automatic Scene Flow Labelling}\label{labelling}

\begin{figure}[!tbp]
    \centering
    \includegraphics[width=0.65\textwidth]{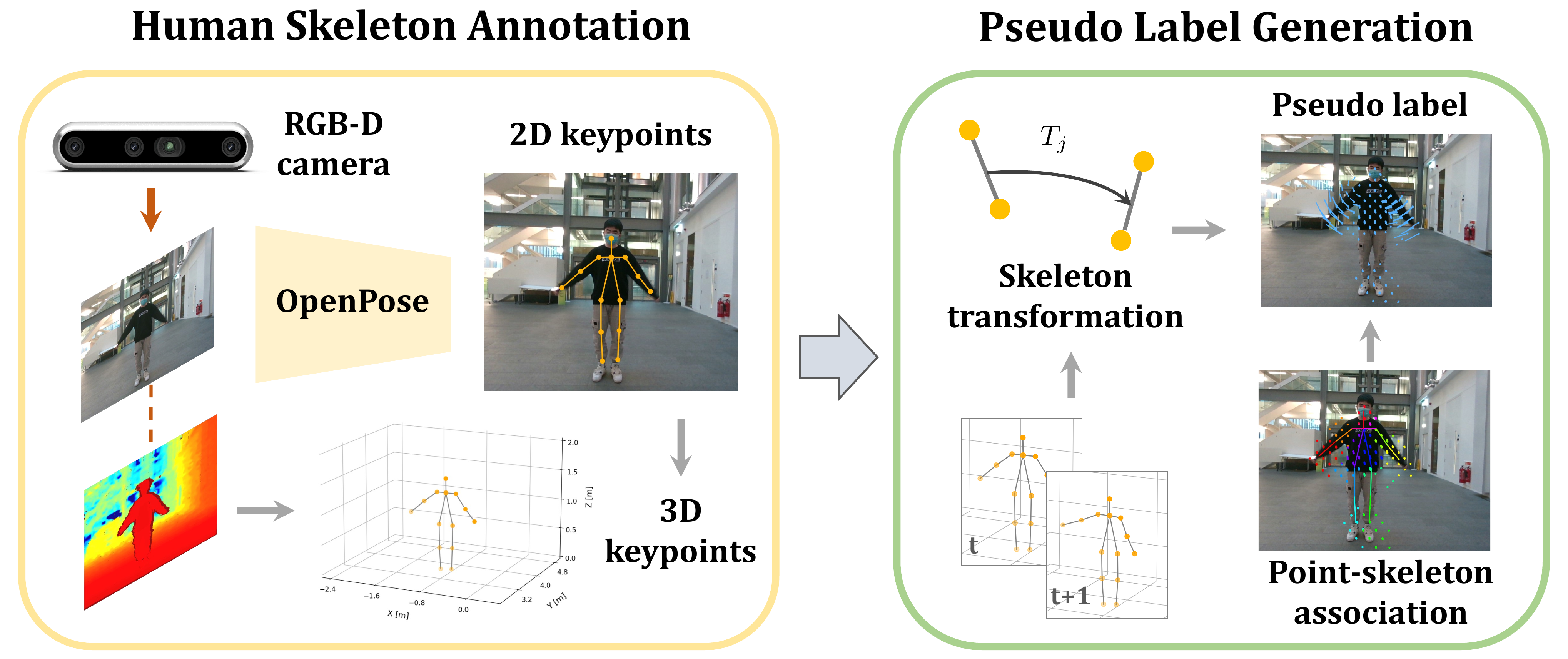}
    \caption{\small{Automatic scene flow labelling pipeline. With the help of the co-located RGB-D camera, we first label 3D human skeletons and then generate noisy pseudo scene flow labels with respect to the skeleton-based rigid-motion assumption. }}
    \label{fig:label}
\end{figure}

To address the human sensing scenarios where non-rigid motion dominates, we propose a cross-modal automatic scene flow labelling scheme (\cf Fig.~\ref{fig:label}), where pseudo scene flow labels are obtained from the 3D human skeletons. Our motivation is based on the observation that the non-rigid human body can be roughly segmented into multiple skeletons each of which can be seen as a rigid body, such as the neck and thigh bone. To simplify the problem of modelling human body motion, we attribute the human dynamics to the rigid motion of these skeletons and further assume the scene flow of points in the vicinity is induced by them.

\noindent\textbf{Human Skeleton Annotation.} 
In human sensing applications, the human skeleton is usually characterised by the 3D position of its two endpoints, aka. keypoints. To quickly and conveniently annotate such keypoints, we refer to the RGB-D images recorded by the co-located RGB-D camera in this work. 
Specifically, we first utilize an open-source pose estimation library (e.g. OpenPose~\cite{cao2019openpose}) to label 2D keypoints on RGB images and then uplift each 2D keypoint to 3D using its corresponding depth value. Then the human skeleton labels can be obtained using intrinsically connected 3D keypoints, as shown in Fig.~\ref{fig:label}.

\noindent\textbf{Pseudo Label Generation.} Given the 3D skeletons automatically annotated above, we can generate pseudo scene flow labels $\bar{\mathcal{F}}=\{\bar{f_i}\in\mathbb{R}^3\}^{N}_{i=1}$ for radar point clouds. For two consecutive point clouds $\mathcal{P}$ and $\mathcal{Q}$, we first compute the inter-frame transformation matrix for all human skeletons in the source frame. Then we assign each source radar point $p_i$ to its closest skeleton and form the point-skeleton association as exhibited in Fig.~\ref{fig:label}. In the final, we derive the pseudo scene flow labels for each selected radar point as $\bar{f_i} = (T_j \circ p_i) - p_i$, where $T_j\in$ is the transformation matrix for the skeleton that point $p_i$ is assigned to. $\circ$ is the action that operates homogeneous transformation for 3D points.


\noindent\textbf{Keypoint-Based Label Filtering.} 
To reduce noise from the inaccuracies of 2D keypoint estimation, we filter pseudo scene flow labels using confidence scores and 3D displacement of keypoints. Keypoints with confidence below 0.5 are first removed, and those with displacement over 0.5m between frames are further filtered out. This process yields a set of valid keypoints per frame, and skeletons with both two end keypoints valid are considered valid. We then generate a mask $\mathcal{M}$,  used below to minimise the impact of noise.

\subsection{Loss Function}\label{loss}
We use the pseudo scene flow labels $\mathcal{\bar{F}}$ to supervise our scene flow network. To mitigate noise, the valid mask $\mathcal{M}$ is used to filter out noisy labels during loss calculation. Initially, our network tended to converge to local minima, producing small and uniform scene flow vectors due to the predominance of small-scale movements in the data. To counteract this, we introduced a weighted loss function that emphasizes points with large-scale movements more significantly. Our adjusted loss is:
\begin{equation}
\mathcal{L} = \alpha_l \mathcal{L}_{large} + \alpha_s \mathcal{L}_{small}
\end{equation}
where $\alpha_l$ and $\alpha_s$ are hyperparameters that balance large- (\ie, larger than a threshold $\zeta$) and small-scale movement impacts, with the $L_2$ distance measuring prediction errors. This approach aims to prevent the network from settling into local minima by reducing the influence of static or minor-moving points.

\section{Experiment}

\subsection{Dataset Collection}
To facilitate the evaluation of our approach, we collect a large-scale multi-modal human motion sensing dataset with annotation labels for various tasks.

\begin{figure}[tbp!]
\centering
\includegraphics[width=\textwidth]{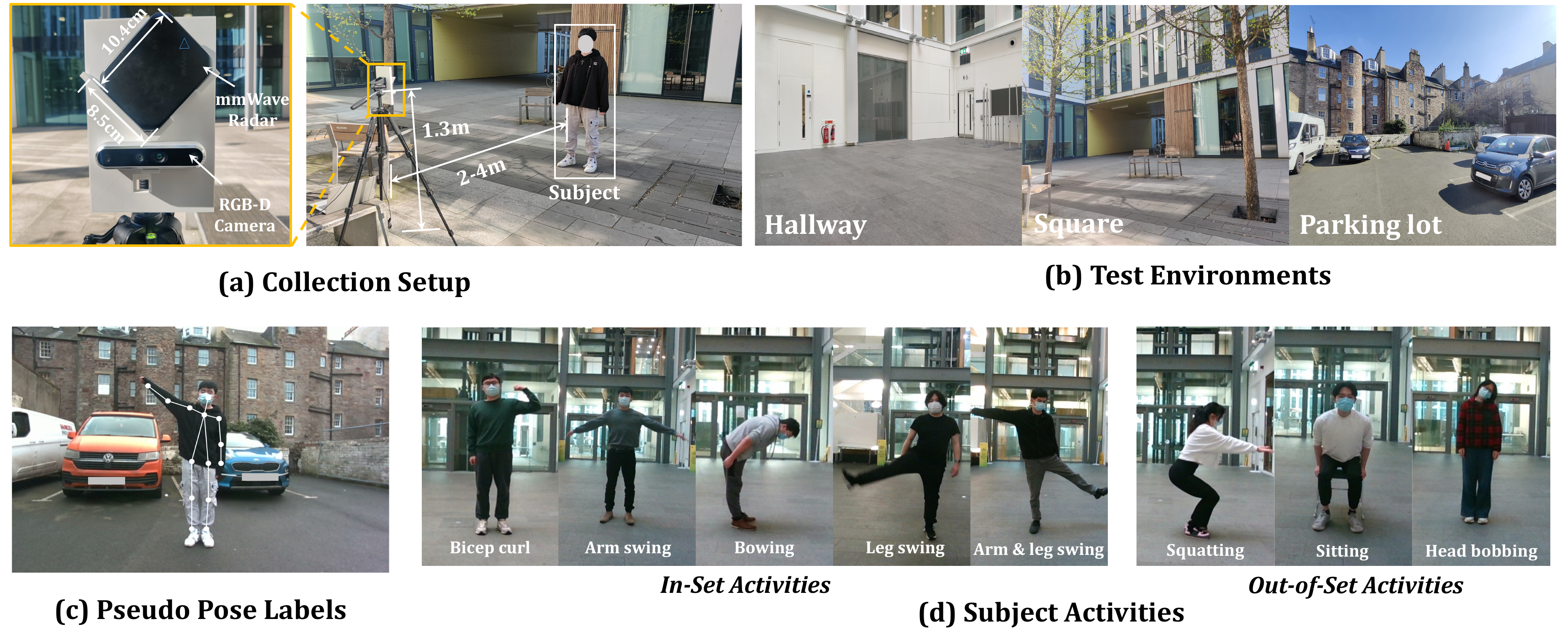}
\caption{\small{Collection setup, test environment, subject activities and pseudo pose labels.}}
\label{fig:setup}
\end{figure}

\noindent\textbf{Platform.} As exhibited in Fig.~\ref{fig:setup} (a), we use a commercial Vayyar vTrigB imaging mmWave radar~\cite{vayyarhome} and a RealSense D455 depth camera~\cite{d455} to capture mmWave radar point clouds and RGB-D images respectively.
Both sensors are fixed on a collection board mounted on a tripod to ensure their relative position is unchanged once calibrated. We alternately query the data frame from two sensors and store them in a PC, resulting in synchronized multi-modal data. Note that the Vayyar radar we used is bespoken designed for fine-grained human sensing. It provides radar data with a higher resolution than most other mmWave radars, \eg, TI radars~\cite{TIMmWaveRadar2024}, thus enabling us to estimate fine-grained scene flow for radar points. 


\noindent\textbf{Procedure.}    
12 participants of various genders, ages and heights are recruited for data collection in this work\footnote{The study has received the ethical approval from the University of Edinburgh, and participant consent forms were signed
before the collection.}. We choose three sites as our human sensing experiment scenes, as shown in Fig.~\ref{fig:setup} (b). Each participant is asked to perform 5 `in-set' activities for each scene, and 3 `out-of-set' activities in one scene for the evaluation of the generalization to unseen activities (\cf Fig.~\ref{fig:setup} (d)). Every subject is asked to wear a face mask to protect their identities from being recognised. The distance between subjects and sensors is 2-4m and randomly selected by the participant during collection. 

\noindent\textbf{Statistics.} After data collection, we crop each sequence to 200 frames for uniform activity distribution. The whole `in-set' dataset consists of $12 \times 3\times 5\times 200 =$ 36k frames in total. To assess the generalization ability to new subjects, we divide the dataset into three parts by subject following the ratio of train:val:test = 3:1:2. The 'out-set' testing set is composed of $12 \times 3\times 200 =$ 7.2k frames.

\subsection{Evaluation Setup}

\noindent\textbf{Data Labelling.}
Our automatic labelling scheme (\cf Sec.~\ref{labelling}) is used to annotate scene flow for the training and validation set by retaining 14 keypoints from OpenPose~\cite{cao2019openpose}, leading to 13 connections between keypoints (\cf Fig.~\ref{fig:setup} (c)). Invalid skeletons are filtered out, and pseudo scene flow labels are assigned to points on valid skeletons. For the testing set, the same pipeline is used, but all points are annotated, with manual inspection and correction to 2D keypoints. For human activity recognition, activities are manually labelled during recording. Human parsing labels are derived from point-skeleton affiliations in scene flow labelling (\cf~\cref{fig:label}), with the valid mask $\mathcal{M}$ applied to filter invalid points during training. The human body part tracking labels are directly obtained from the 3D keypoints identified in the scene flow labelling process.

\noindent\textbf{Evaluation Metric.} For scene flow evaluation, we use the EPE3D (m) and Acc3D metrics following~\cite{baur2021slim,ding2022raflow}. Specifically, we redefined the Acc3D strict and relax requirements by reducing 0.05/0.1m to 0.025/0.05m to adapt to our human-centric scenario. The overall accuracy (OA) (\%) is reported for both the human activity recognition and human parsing tasks, while mIoU (\%) is used for HP only. To evaluate the human body part tracking, we compute the mean joint localization error (mJE) (m).

\begin{table}[t!]
\centering
\begin{minipage}[t]{0.475\linewidth} 
    \renewcommand\arraystretch{1.0}
    \setlength\tabcolsep{2pt}
    \centering
    \caption{{Comparison of scene flow results between ours and state of the arts. $\uparrow$ means bigger values are better while $\downarrow$ means smaller values are better.}}
    \resizebox{\columnwidth}{!}{%
    \begin{tabular}{@{}lcccccc@{}}
    \toprule
        & \multicolumn{3}{c}{EPE3D (m) $\downarrow$} & \multicolumn{2}{c}{Acc3D $\uparrow$}\\
        \cmidrule(r){2-4} \cmidrule(r){5-6}
        Method & ~All~ & ~Moving~ & ~Static~ &  Strict  & Relax    \\
    \midrule 
      FlowNet3D~\cite{liu2019flownet3d} & 0.293 & 0.290 & 	0.259 & 0.016 &	0.095\\
      PPWC-Net~\cite{wu2020pointpwc} & 0.171 & 0.181 & 0.128	& 0.138	& 0.179\\
      Graph Prior~\cite{pontes2020scene} & 0.315 & 	0.322 & 0.283 & 0.007 & 0.011\\
      FLOT~\cite{puy2020flot} & 0.299	& 0.307	& 0.265 & 0.015	& 0.094 \\
      FlowStep3D~\cite{kittenplon2021flowstep3d} & 0.243 & 0.251 & 0.216 & 0.062 & 0.109\\
      NSFP~\cite{li2021neural} & 0.197 & 0.213 & 0.167 & 0.085 & 0.143 \\
      PV-RAFT~\cite{wei2021pv} & 0.161 & 0.170	& 0.107 & 0.179	& 0.292 \\
      RaFlow~\cite{ding2022raflow} & 0.107 & 0.115 &	0.094 & 0.271 & 0.427 \\
      Bi-PFNet~\cite{cheng2022bi} & 0.159	& 0.168	& 0.111	& 0.153 & 0.264\\
      milliFlow (ours) & \textbf{0.046} &	\textbf{0.051} &	\textbf{0.009} &	 \textbf{0.406} & \textbf{0.703}\\
    \bottomrule
    \end{tabular}
    }
    \vspace{-2em}
    \label{tab:baselines}
\end{minipage}
\hfill 
\begin{minipage}[t]{0.475\linewidth} 
    \begin{minipage}[t]{\linewidth}
    \renewcommand\arraystretch{1.0}
    \setlength\tabcolsep{3pt}
    \scriptsize
    \centering
    \caption{\small{Breakdown results of our scene flow network. }}
    \vspace{-1em}
    \resizebox{\columnwidth}{!}{%
    \begin{tabular}{@{}l|lcccccccc@{}}
    \toprule
        & &  \multicolumn{3}{c}{EPE3D (m) $\downarrow$} & \multicolumn{2}{c}{Acc3D $\uparrow$}\\
        \cmidrule(r){3-5} \cmidrule(r){6-7}
        & Method & ~All~ & ~Moving~ & ~Static~ &  Strict & Relax    \\ 
    \midrule 
    (a) & Full version & \textbf{0.046} &	\textbf{0.051} &	\textbf{0.009} &	 \textbf{0.406} & \textbf{0.703}\\
    (b) & (a) w/o TP & 0.053 & 0.062 & 0.018 & 0.382 & 0.676\\
    (c) & (b) w/o GA & 0.061 & 0.068 & 0.025 & 0.361 & 0.628\\
    (d) & (c) w/o CF & 0.071 & 0.077 & 0.028 & 0.315 & 0.536\\
    (e) & (d) w/o CR & 0.083 & 0.090 & 0.034 & 0.286 & 0.490\\
    \bottomrule
    \end{tabular}
    }
    \label{tab:breakdown}
    \end{minipage}
    \begin{minipage}[t]{\linewidth}
    \renewcommand\arraystretch{1.0}
    \setlength\tabcolsep{6pt}
    \centering
    \caption{ 
    \small{Generalization of our model to new activities.}}
    \vspace{-1em}
    \resizebox{\columnwidth}{!}{%
    \begin{tabular}{@{}lcccccc@{}}
    \toprule
        & \multicolumn{3}{c}{EPE3D (m) $\downarrow$} & \multicolumn{2}{c}{Acc3D $\uparrow$}\\
        \cmidrule(r){2-4} \cmidrule(r){5-6}
        Activity & ~All~ & ~Moving~ & ~Static~ &  Strict & Relax    \\
    \midrule 
      Sitting & 0.034	& 0.038	& 0.004	& 0.498	& 0.771\\
      Squatting & 0.040 & 0.047 & 0.009 & 0.416 &	0.688\\
      Head bobbing & 0.027 & 0.031 & 0.006 &	0.664 & 0.859\\
    \midrule 
      Average & 0.034	& 0.039	& 0.006	& 0.526 &	0.773\\
    \bottomrule
    \end{tabular}
    }
    \label{tab:out-of-set}
    \end{minipage}
\end{minipage}
\end{table}

\subsection{Scene Flow Evaluation}
\noindent\textbf{Compared to the State of the Arts.} 
We first compare our scene flow network with state-of-the-art methods of point-based scene flow estimation. Our baselines include 7 deep learning-based methods~\cite{liu2019flownet3d,wu2020pointpwc,kittenplon2021flowstep3d,puy2020flot,wei2021pv,cheng2022bi,ding2022raflow} whose networks are also supervised with pseudo scene flow labels, and two non-learning-based method Graph Prior~\cite{pontes2020scene} and NSFP~\cite{li2021neural} that solves scene flow via online optimization. The evaluation results on the `in-set' testing set can be seen in Table~\ref{tab:baselines}.
The scene flow network of milliFlow achieves a cm-level average EPE3D (\ie, 4.6cm) and a high relax Acc3D of 70.3\%, ranking 1st on all metrics with a large margin compared to the baselines. This satisfactory performance proves the efficacy of our method to address the challenges existing in our scene flow task. Note that our network is trained with pseudo labels automatically generated using RGB-D images (\cf Sec.~\ref{labelling}), which does not demand any manual annotation efforts. 

\begin{figure*}[tbp!]
    \centering
    \includegraphics[width=\textwidth]{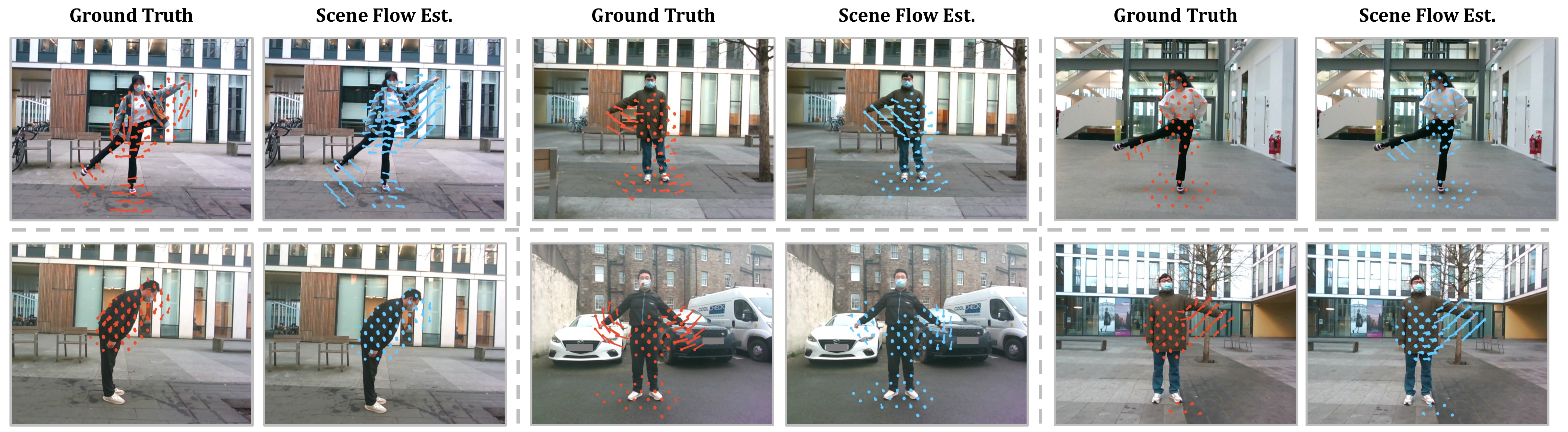}
    \caption{\small{Qualitative scene flow results. radar points and scene flow vectors are projected onto the image and the colour \textcolor{flowred}{red} is used for the ground truth while \textcolor{flowblue}{blue} for ours. }}
    \label{fig:qual}
\end{figure*}

\noindent\textbf{Qualitative Results.} 
Our output examples, illustrated in Fig.~\ref{fig:qual}, effectively demonstrate our method's ability to produce reliable scene flow for diverse activities and subjects across three environments. Despite the inherent sparsity and noise in radar point clouds, our network adeptly learns representative features through global feature aggregation and robust scene flow vector regression. Notably, even when some body parts are absent in the radar data, our approach compensates by utilizing historical information, thereby benefiting the current scene flow estimation and ensuring robust performance across successive frames.


\noindent\textbf{Runtime Efficiency.} We test the runtime efficiency of our scene flow network on a single NVIDIA RTX 3090 GPU. Given sequential testing radar point clouds, we feed them one by one into our trained model for inference. As a result, our model has real-time performance with one inference step in 74ms ($\sim$13.5Hz). Moreover, the maximum allocated GPU memory is only 134 MB during inference. This minimal memory usage enables our network to operate in parallel with downstream networks, underscoring its practicality for real-time applications.

\noindent\textbf{Ablation Study.} 
To validate the effectiveness of key components within our scene flow network, we systematically disabled each, \ie, temporal propagation (TP), global aggregation (GA), context feature (CF), and constrained regression (CR), and observed their effects on performance, as presented in Table~\ref{tab:breakdown}.  Overall, the full version of our network (row (a)) yields the best results and each component helps to elevate the performance on each metric (from row (e) to (a)).  By utilizing information from previous frames, temporal propagation significantly improved scene flow estimation, achieving a 13.2\% reduction in EPE3D and a 6.3\% increase in strict Acc3D accuracy. Global aggregation, employing an attention mechanism, bolstered local features with global context, leading to a 13.1\% improvement in EPE3D. The inclusion of context features into our flow embedding, aimed at retaining source frame context, provided a modest but surprising boost, lowering the overall EPE3D by 0.01m. Lastly, the biggest improvement (\ie, a 0.012m decrease on EPE3D) is brought by our constrained regression, in which the scene flow component on each axis is clamped by a fixed threshold (\eg 0.1m). This is reasonable as non-viable results, for example, a scene flow vector with a length of 0.5m, can seriously degrade our results.

\noindent\textbf{Generalization to New Activities.}
In the above experiments, we evaluate our method on the testing set whose subjects are unseen during training. The results demonstrate the generalization ability of our trained model to new users that perform the same `in-set' activities. To further test its generalization to new activities, here we evaluate our trained model on the `out-of-set' testing set in which three performed activities are not included in the training set. We can see from Tab.~\ref{tab:out-of-set} that, our trained model can still keep an equally good performance when encountering unseen activities. This demonstrates the ability of our model to cope with new activities in human sensing. We also observed that the performance to new activities is better than the overall performance shown in~\cref{tab:baselines}. This is because our subjects either keep most of their bodies static (\ie, heading bobbing) or stay completely static in many frames (\ie, sitting and squatting) when doing unseen activities. As a result, estimating the scene flow of points belonging to them is much easier. We believe that our trained model can also generalize to other daily human activities.


\subsection{Downstream Task Evaluation}\label{downstream_eval}

\begin{table}[t!]
\centering
\begin{minipage}[t]{0.49\linewidth} 
    \renewcommand\arraystretch{1.0}
    \setlength\tabcolsep{9pt}
    \centering
    \caption{\small{Evaluation on the benefit of scene flow for the HAR task. }}
    \resizebox{\columnwidth}{!}{%
    \begin{tabular}{@{}lcccccccccc@{}}
    \toprule
         Method &  Raw & w. S1 & Gain &  w. S2 & Gain \\
    \midrule
      Ours & 47.32 &   57.88 & +10.56 & 57.78 & +10.46\\
      MMPointGNN~\cite{9533989} & 52.46 & 60.16 & +7.70 & 59.94 & +7.48\\ 
      RadHAR~\cite{singh2019radhar} & 44.65 & 49.98 & +5.33 & 50.53 & +5.88\\
      \midrule
      Average & 48.14 & 56.01 & +7.87 & 56.08 & + 7.94\\
    \bottomrule
    \end{tabular}
    }
    \label{tab:har}
\end{minipage}
\hspace{0.5cm} 
\begin{minipage}[t]{0.44\linewidth} 
    \renewcommand\arraystretch{1.0}
    \setlength\tabcolsep{4pt}
    \centering
    \caption{\small{Evaluation on the benefit of scene flow for the HP task. }}
    \resizebox{\columnwidth}{!}{%
    \begin{tabular}{@{}lcccccccccc@{}}
    \toprule
         Method  & mIoU (\%) & Gain (\%) & oA (\%) & Gain (\%) \\
    \midrule
      Raw  &  49.09 & - & 65.75 & -\\
      w. S1 & 52.72 & +3.63 & 69.27 & +3.52\\
      w. S2 & 51.04 & +1.95 & 68.21 & +2.46\\
    \bottomrule
    \end{tabular}
    }
    \label{tab:hp}
\end{minipage}
\end{table}
As a low-level signal in understanding motions, scene flow can directly enhance low-quality radar point clouds with full per-point displacement information between two frames. Moreover, the latent spatial-temporal representations can be implicitly learned by guiding the network to estimate scene flow, which can be used to support other tasks. Therefore, we envision that learning scene flow estimation can benefit a wide range of higher-level downstream tasks in mmWave-based human sensing. To demonstrate the benefit, here we consider three representative downstream tasks, including human activity recognition (HAR), human parsing (HP) and human body part tracking (HBPT) for evaluation. 


\noindent\textbf{HAR Evaluation Setup.} HAR plays a significant role in a wide range of applications, such as elderly healthcare monitoring~\cite{taylor2020intelligent}, smart home~\cite{bianchi2019iot} and behaviour surveillance~\cite{khurana2018deep}. To validate the functionality of scene flow to benefit HAR, we design a HAR base network by combining some key components from~\cref{network} and also utilizing the LSTM network~\cite{hochreiter1997long} to track the temporal relationship before regressing the classification scores. Besides, two state-of-the-art methods~\cite{9533989, singh2019radhar} bespoken for mmWave-based HAR are selected for a more persuasive comparison. Particularly, we design two strategies to harness the scene flow network as a plug-and-play module to the HAR network. Strategy 1 (S1) directly takes the estimated scene flow as point-level raw features and decorates each radar point with them, while Strategy 2 (S2) leverages the latent representations encoded by scene flow networks to enhance the low-quality radar data. 

\begin{figure}[t!]
\centering
\begin{minipage}[t]{0.42\linewidth} 
  \centering
  \includegraphics[scale=0.135]{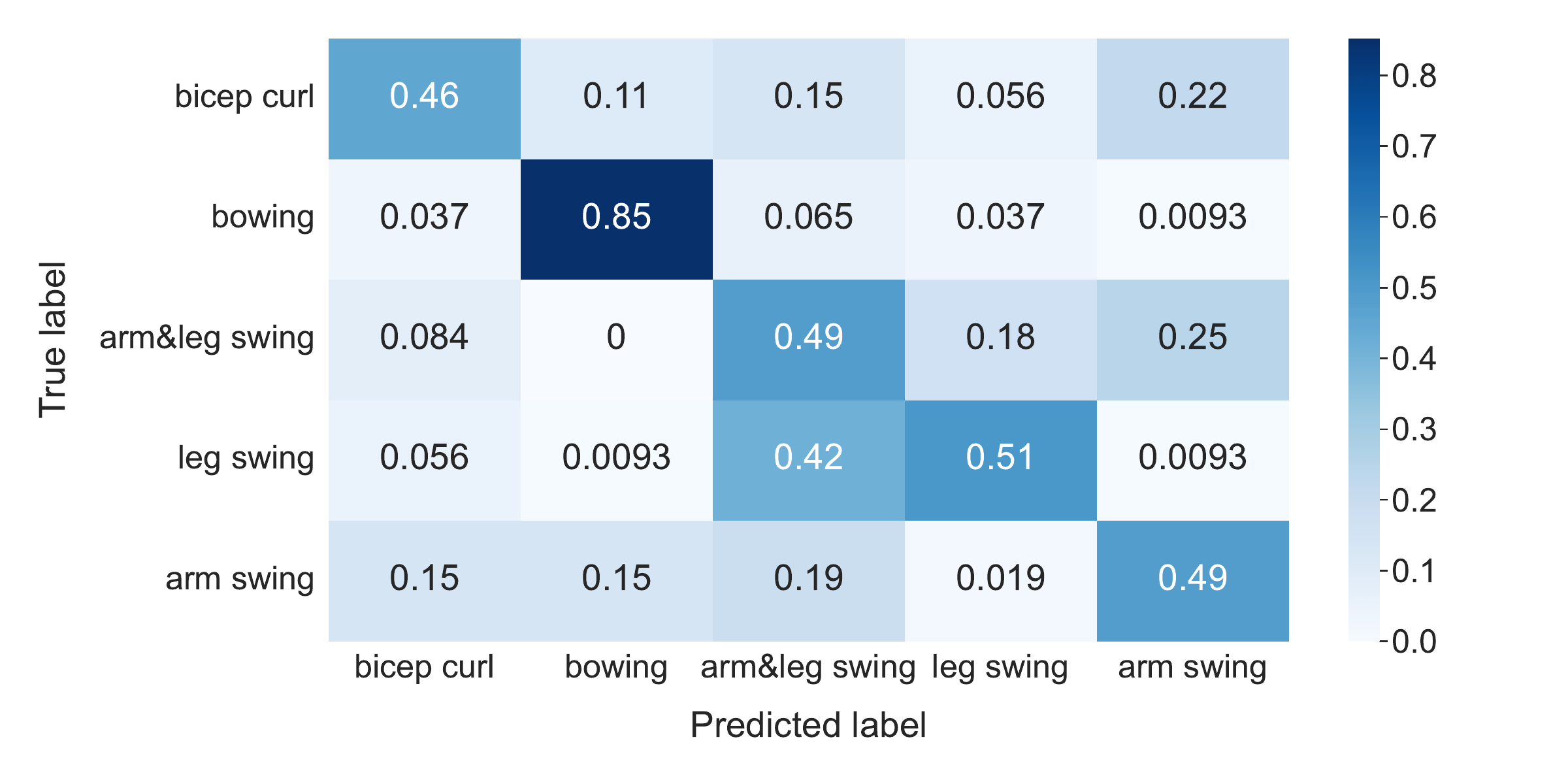}
  \caption{\small{Confusion matrix of HAR. }}
  \label{fig:confusion}
\end{minipage}
\hspace{0.5cm} 
\begin{minipage}[t]{0.46\linewidth} 
    \renewcommand\arraystretch{1.0}
    \setlength\tabcolsep{12pt}
    \vspace{-8em}
    \captionof{table}{
    \small{Scene flow-based HBPT evaluation results.}}
    \centering
    \resizebox{\columnwidth}{!}{%
    \begin{tabular}{@{}lcccccc@{}}
    \toprule
        & \multicolumn{4}{c}{Tracking length - mJE (m) $\downarrow$} \\
        \cmidrule(r){2-5} 
        Activity & 1 & 2 & 3 & 4 \\
    \midrule 
     Arm swing  & 0.028	& 0.076	& 0.097	& 0.124	\\
     Leg swing  & 0.016	& 0.071	& 0.105	& 0.130\\
     Arm \& leg swing & 0.030& 0.108	& 0.146	& 0.178\\
     \midrule
     Average & 0.025 & 0.085 & 0.116 & 0.144 \\
    \bottomrule
    \end{tabular}
    }
    \label{tab:hbpt}
\end{minipage}
\end{figure}

\noindent\textbf{HAR Evaluation Results.} We evaluate our proposed two strategies for scene flow application to HAR task on our `in-set' testing set. The length of each input sequence $T$ for HAR is set as 20. The evaluation results are shown in Table~\ref{tab:har}. As we can see, both two proposed strategies can enhance the performance of our base network and state-of-the-art methods on HAR,
demonstrating their usefulness in helping distinguish between different human activities. It can be also observed that the average accuracy shown in~\cref{tab:har} is relatively low compared to those reported in~\cite{9533989, singh2019radhar}. We credit this to our subject activities being very similar and inter-included to some extent. For example, the arm \& leg swing activity can be separated into arm swing and leg swing, which are also inside our `in-set' activities.  As a result, both ours and the two states of the arts find it more difficult to correctly distinguish between these activities. The breakdown results (ours with S1) on HAR as the confusion matrix is shown in Fig.~\ref{fig:confusion}. As we can see, our HAR network achieves a relatively high accuracy value on the bowing activity since it has apparently different motion patterns from others.  However, the accuracy values on other activities are not as satisfactory as on bowing. For example, 42\% leg swing samples are wrongly predicted as the arm \& leg swing activity. This further justifies our explanation that some activities are so similar and thus can easily confuse our HAR networks in classification. 

\begin{figure}[tp!]
  \centering
  \includegraphics[scale=0.155]{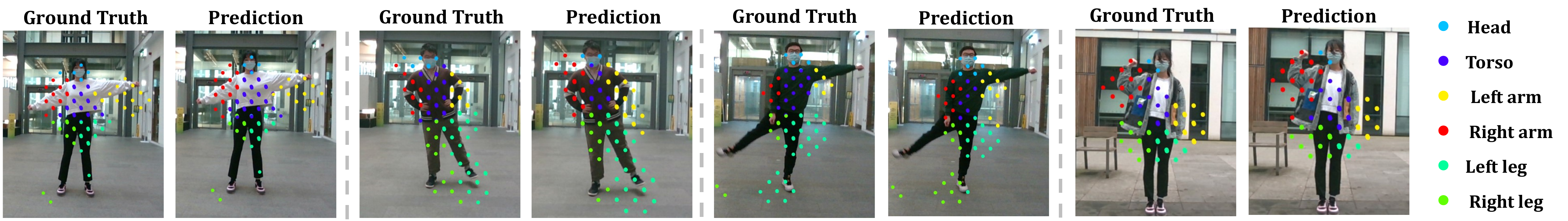}
  \caption{\small{Human parsing results visualization.}}
  \label{fig:parsing}
\end{figure}

\noindent\textbf{HP Evaluation}
Human parsing aims to parse human semantic body segments (head, arms, torso, etc.) from sensor data. With radar point clouds as input, our objective of HP is to identify the body parts that correspond to each point. To evaluate the effectiveness of scene flow on this task, we also designed a base network for this downstream task using key components from~\cref{network}. Both two scene flow application strategies proposed for the HAR above are also used here for the HP task. The number of body segments for parsing is 6, encompassing two arms, two legs, head and torso. As seen in~\cref{tab:hp}, leveraging scene flow in two strategies also contributes to better performance on the HP task.
Directly applying scene flow to points (S1) yields better results than using latent feature recycling (S2), as the former provides explicit per-point motion information. In contrast, latent features are less direct and require additional processing. Some examples of our HP results (with Strategy 1)  are exhibited in Fig.~\ref{fig:parsing}. Thanks to the enhancement by the scene flow, our method can accurately parse radar point clouds into different body parts, close to the ground truth results.

\noindent\textbf{HBPT Evaluation.} Given the initial position of a human body part, our human body part tracking task aims to track its movement in subsequent frames. 
With point-level scene flow estimation, we first group $N_t$ radar points that can be assigned to the skeleton needed to be tracked (\cf~\cref{fig:label}). Then we generate natural correspondences $\{p_i, p_i+f_i\}_{i=1}^{N_t}$ using their scene flow and apply the classical Kabsch algorithm~\cite{kabsch1976solution} to solve the rigid skeleton transformation based on them. This updates the endpoints' positions, allowing for tracking of the human body part across frames without additional training, relying only on a pre-trained scene flow network. For evaluation, we select the sequences of arm swing, leg swing and arm \& leg swing activities in the testing set and take the two arms, legs and both of them respectively as our tracking targets.

\noindent For implementation, we divide the long sequences into short clips with a length of 5 frames and track the body parts within each clip. After initializing the ground truth positions in the first frame, we aim to track each skeleton for the next four frames. The HBPT results at different tracking lengths on three activities are reported in Fig.~\ref{tab:hbpt}. With accurate scene flow estimation, we can effectively track multiple body parts together for different activities. For one-frame tracking, our method achieves an average mJE of <3cm on three activities, demonstrating the capability of scene flow to enable the HBPT task. However, the errors become larger as the tracking length increases, which indicates the happening of the tracking drift. This is inevitable for our method as more radar points associated with other skeletons will be wrongly induced for transformation calculation when the tracking continues. We also observe that the performance on the arm \& leg swing activity is worse than the other two activities. This is reasonable as we need to track twice the skeletons in this activity as others, where more skeletons may interfere with each other during tracking.


\section{Conclusion}

In this paper, we introduce \textit{milliFlow}, a deep learning framework designed to estimate scene flow for enhancing 3D mmWave radar point clouds in human motion sensing. Addressing mmWave's inherent instability and sparsity, \textit{milliFlow} integrates multi-scale local and global features with temporal data. We also develop an automated labeling method to reduce the need for costly manual annotation. Extensive testing on our dataset and three downstream tasks shows \textit{milliFlow}'s effectiveness, suggesting its potential as a valuable component in human motion sensing systems, significantly improving their performance.

\section*{Acknowledgements}

This research is partially supported by the Engineering and Physical Sciences Research Council (EPSRC) under the Centre for Doctoral Training in Robotics and Autonomous Systems at the Edinburgh Centre of Robotics (EP/S023208/1).

%
%
\appendix

\section*{Appendix}

\vspace{1em}
The appendix is organized as follows:
\begin{itemize}[leftmargin=*,label=$\bullet$]
    \setlength{\itemsep}{0pt}
    \setlength{\parsep}{0pt}
    \setlength{\parskip}{0pt}
    \item \cref{preliminary} illustrates more details about the mmWave radar we used for experiments and two deep learning layers for point cloud processing.
    \item \cref{implementation} presents more implementation details on sensor specifications, data preprocessing, evaluation metrics and network and application strategies. 
    \item \cref{analysis} provides a sensitivity analysis of our scene flow network against aggregation mechanisms, test environment and the number of training subjects.
    \item \cref{limit} explains several limitations to be considered for future work.
\end{itemize}

\section{Preliminary}\label{preliminary}

\subsection{SFCW mmWave Radar}\label{pc_generation}
In this work, we employ the Vayyar radar~\cite{vayyarhome}, which is a stepped-frequency continuous-wave (SFCW) radar~\cite{weiss2009continuous} bespoken designed for fine-grained human sensing. This subsection aims to provide a concise overview of the point cloud generation principles of SFCW radar to aid in comprehending the hardware components of the proposed system. In what follows we introduce the working principle of SFCW radars.

\noindent\textbf{Range Estimation.}
With SFCW radar, we can detect range information with a single Tx-Rx pair. For a CW radar, the transmitting signal can be written as $s(t)=Asin(2\pi f_0 t)$, and the corresponding receiving signal from the reflection on an object at range $R$ would be $s_r(t)=A_rsin(2\pi f_0t - \phi)$, where $\phi$ is the phase difference of the transmitting and receiving signals, and can be written as $\phi = 2\pi f_0 \frac{2R}{c}$. As a result, we are able to estimate the range as $R=\frac{c}{4\pi f_0}\phi$. Note that $\phi$ can only be in range $[0, 2\pi]$. As a result, for CW radar that works on a single frequency, the maximum unambiguous range is very limited. For example, if $f_0 = 2GHz$, $R_{max} = 7.5cm$. With SFCW radar, we are able to get the phase difference at different frequencies, which provides much richer information for range estimation. For two consecutive frequencies $f_1$ and $f_2$, the frequency difference $\Delta f = f_2 - f_1$, and the phase differences at each frequency step are $\phi_1 = \frac{4\pi R}{c}f_1$ and $\phi_2 = \frac{4\pi R}{c}f_2$. We have the following equation:
\begin{equation}
    \Delta \phi = \phi_2 - \phi_1 = \frac{4 \pi R}{c}(f_2 - f_1) = \frac{4\pi R}{c} \Delta f
\end{equation}
and we get
\begin{equation}
    R=\frac{c}{4\pi \Delta f} \Delta \phi
\end{equation}
Also, we have $\Delta \phi \in [0, 2\pi]$, so the maximum unambiguous range is inversely proportional to $\Delta f$, which is the frequency step in SFCW radar. 
During each frame, the transmitter sequentially emits a series of waves of increasing frequency at equal intervals. The transmitted signal is reflected back at different surfaces in the scene and received by the receiving antenna. The ADC samples the change in amplitude and phase between the transmitted signal and the received signal and stores the values in IQ format. We can apply the Fourier Transform to get the phase variation, with which we can further derive the range of the objects. This is also known as `Range-FFT'. The range resolution is inversely proportional to the bandwidth of the radar, written as $R_{res} = \frac{c}{2B}$, where $B = f_{max} - f_{min}$. 


\noindent\textbf{Angle-of-arrival Estimation.}
Angle-of-Arrival (AoA) estimation can be performed using a linear receiver array with equally-spaced elements, typically separated by a distance of $1/2 \lambda$, where the number of elements in the array, denoted as $N$, is greater than or equal to two. When a reflected signal is received at each element of the array, a phase difference arises due to the slight differences in signal travel distance, as determined by the angle of arrival, $\theta$. This phase difference can be quantified using the formula $\Delta \phi = kdsin(\theta)$, where $k$ represents the wave number and $d$ is the distance between consecutive elements. The angle of arrival can then be estimated by analyzing the phase difference using a Fourier Transform technique, commonly referred to as `Angle-FFT'.
Vayyar uses Multiple-Input Multiple-Output (MIMO) array, and for each frame, we are able to get a 2D virtual antenna matrix, from which we can estimate azimuth and elevation AoA simultaneously. 

\noindent\textbf{Point Cloud Generation.}
For each frame, the raw data is stored as a complex matrix of size $N_a * N_e * M$, where $N_a$ and $N_e$ are the numbers of Tx-Rx antenna pairs (i.e., virtual antennas) in azimuth and elevation directions, respectively, and M is the number of frequency steps. First, the clutter removal is applied to focus on dynamic objects. We then apply Range-FFT for Range estimation for each virtual antenna and further perform 2D Angle-FFT to get the AoA information in both directions. Following the above processing, we are able to get a 3D heatmap for each frame.
Strong peaks are then detected with methods like CFAR~\cite{scharf1991statistical}, and converted to 3D points, which is the input to our scene flow estimation. 

To summarize, given the 3D data cube, mmWave radar conducts the range-FFT, Doppler-FFT and two `Angle-FFT' to obtain a 4D radar heatmap. Strong peaks are then detected to generate a radar point cloud, which is the input to our scene flow estimation model. 

\subsection{Deep Learning Layers for Point Cloud}

\subsubsection{Set Abstraction Layer.}\label{sa_layer}
In our radar scene flow network, we adopt the set abstraction layer proposed in PointNet++~\cite{qi2017pointnet++} as the basic learning layer to extract local point features. As an extension of PointNet~\cite{qi2017pointnet}, PointNet++~\cite{qi2017pointnet++} proposes a hierarchical structure composed of multiple set abstraction levels. At each level, the local features of a set of points are abstracted and taken as the input to the next level. The working process of each set abstraction layer can be summarized in three steps. First, the iterative farthest point sampling (FPS) is applied to select $N'$ points the point sets as the centroids for local regions. Second, the ball query identifies neighbour points within a radius $R$ of each centroid, resulting in $N'$ local point sets.  Lastly, the local region features are encoded for each centroid using the PointNet layer~\cite{qi2017pointnet}, which is composed of an MLP to extract high-level representations and a max-pooling layer to aggregate the representations of all local points to the centroid. The output is the per-point local features for $N'$ selected points at a specific scale determined by the radius $R$. We believe that this learning layer can effectively gather information from neighbour points to each radar target.

\subsubsection{Cost Volume Layer.}\label{cv_layer}
To encode points motion between two radar frames, in our network design, we leverage the point cloud cost volume layer in~\cite{wu2020pointpwc} to correlate features. In the cost volume layer, the matching costs $Cost(p_i, q_j)$ of all point pairs $(p_i, q_j)$ between two point clouds $\mathcal{P}$ and $\mathcal{Q}$ are first calculated through an MLP. Then, the costs are aggregated in a patch-to-patch manner to produce robust and stable cost volumes. During aggregation, the neighbourhood is first found for each point $p_c$ in $\mathcal{P}$. Then, for each neighbour point $p_i$ of $p_c$, a neighbourhood is found around it in $\mathcal{Q}$. With these patches in $\mathcal{Q}$ and $\mathcal{P}$, the costs can be aggregated respectively through a weight-sum operation whose weights are leanrable parameters from MLPs. The output is two-dimension cost volumes with shape $N\times D$, where $N$ is the number of points of the point cloud $\mathcal{P}$ and $D$ is the dimension of the cost volume.
\section{Implementation Details.}\label{implementation}

\subsection{Sensor Specifications}
As said in the main paper, we use a commercial Vayyar vTrigB imaging mmWave radar~\cite{vayyarhome} and a RealSense D455 depth camera~\cite{d455} to capture mmWave radar point clouds and RGB-D images respectively. The mmWave radar device is portable with a size of 10.4 cm $\times$ 8.5cm and a weight of 110g in total, which is designed following the stepped frequency continuous wave (SFCW) principles (\cf~\cref{preliminary}). With the default sensor setting, the range resolution of our mmWave radar is 9.35 cm, and the maximum range is 14 m, while the angular resolution is approximately 6.7 degrees.
The frequency bandwidth is set as 1.6GHz (from 62 to 63.6 GHz) and the frequency step is 10.66 MHz, which results in 151 frequency samples for each frame. 16 RX and 20 Tx antennas on its PCB are employed for producing radar data, which equals 320 Rx/Tx virtual pairs. 
The internal data processing pipeline of this device follows the steps we illustrate in Sec.~\ref{pc_generation}. The final output is a set of 3D points with per-point intensity values. For the depth camera, the RGB-D image size is set as 640$\times$480 and its depth measurement range is from 0.6m to 6m. The average frame rate over all collected data is $\sim $13.2Hz. 

\subsection{Data Preprocessing}\label{preprocessing} Given sequences of mmWave radar point clouds, we first filter them by range to discard points outside our region of interest. The side, forward and height range in the radar coordinate frame is set as  [(-3, 3), (0.5, 5), (-1.5, 1.5)] meters respectively. We then filter points by intensity to omit the background points. The intensity threshold is empirically set as 0.5. For all training and validation frames, we randomly sample 128 points from each radar point cloud to facilitate fast mini-batch-based training. We keep the number of points unchanged in all testing frames to make sure that the scene flow estimation for each point can be examined. In the end, we generate scene flow samples by combing pair-wise frames. Each 200-frame sequence can produce 199 samples, each of which consists of two consecutive mmWave radar point clouds and their corresponding RGB-D images. 

\subsection{Evaluation Metrics}

We use the following evaluation metrics to quantify the performance of scene flow estimation and downstream human sensing tasks. 

\begin{itemize}[leftmargin=*]
\setlength{\itemsep}{0pt}
\setlength{\parsep}{0pt}
\setlength{\parskip}{0pt}
\item \textbf{EPE3D (m)}. It computes the average 3D endpoint error $||f_i-f_{gt}||_2$ over all points in a frame. Following~\cite{baur2021slim,ding2022raflow}, we also report the errors separately for the \emph{moving} and \emph{static} points. Points with a ground truth flow vector larger than 0.01m are labelled as \emph{moving}. 
\item \textbf{Acc3D}. It measures the ratio of points in one frame that meets the strict/relax accuracy requirements, i.e., either EPE3D < 0.05/0.1m or relative error <5\%/10\%. 
These two metrics (strict and relax) can represent the estimation accuracy well in autonomous driving scenarios~\cite{wu2020pointpwc,baur2021slim}. However, in our human sensing scenarios, the movement scale of points is often smaller than 0.05m per frame. As a result, almost all points can easily meet the 0.05/0.1m requirement. To better examine and compare different methods, we redefine these two metrics by reducing 0.05/0.1m to 0.025/0.05m while keeping the relative error 5\%/10\% requirement unchanged.

\item \textbf{Overall accuracy (oA) (\%)}. It measures the proportion of examples for which the predicted label matches the single target label, i.e., $\frac{TP+TN}{TP+FP+TN+FN}$. In our experiments, we use it to formulate HAR and HP problems as per-sequence and per-point classification problems.
\item \textbf{Mean Intersection over Union (mIoU) (\%)}. For a single body part class, its IoU is defined as $\frac{TP}{TP+FP+FN}$, measuring the overlap between the predicted subset of points and the ground truth one divided by their union. The mean IoU (mIoU) score is calculated by averaging scores on all body part classes. In our evaluation, we use this metric for our HP tasks to show the paring results.
\item \textbf{Mean Joint Localization Error (mJE) (m)}. This metric is used for our HBPT task. We define it as the mean 3D Euclidean distance between the positions of endpoints of the tracked body parts and their ground truth positions. 
\end{itemize}

\subsection{Network and Training Details}
For the scene flow network, the grouping radii for four SA layers used for local or context feature extraction are [0.05, 0.1, 0.2, 0.4] meters. The numbers of local samples for them are set as [4, 8, 16, 32] and the dimension of the MLP used in and after each SA layer is [32, 32, 64], [64, 64, 64] respectively. In all global feature aggregation, the MLP used for attention mapping is two-layer with a hidden dimension of 128. In the CV layer, the number of neighbours for patch-to-patch aggregation is set as 8 and each MLP used to learn aggregation weights has three layers with hidden dimensions of [8, 8]. The local encoder used to generate flow embedding has the same hyperparameters as the former one except for its MLP which has the dimension of [512, 256, 64] in each SA layer. The dimension of the flow regressor MLP is [256, 128, 64, 3] and the threshold $\epsilon$ to constrain the output is set as $0.1m$. To implement our temporal propagation, we divide the long sequence of scene flow samples into many mini-clips with a length of 5 frames and train using mini-batches of them after shuffling. During inference, we re-initialize the hidden state of the GRU to zero vectors after 5 frames to fit the training pattern. 

The threshold $\zeta$ used for our loss function is fixed as 0.1m for all experiments while the weight hyperparameters $\alpha_l$ and $\alpha_s$ are set as 2 and 1. For all training in our experiments, we consistently use the Adam optimizer with an initial learning rate of 1e-3, which decays by 0.9 after each epoch. The performance on the validation set is used to determine when to stop the training process and to select the best model for each training experiment. 

\subsection{Downstream Network}

\noindent\textbf{HAR Base Network.}
Given a sequence of $T$ radar point clouds together with their point-level features as input, we follow the process in local feature abstraction and global feature aggregation to extract the local-global representations for each point cloud. Then we use another local encoder to extract higher-level features and aggregate the global feature vector again for each frame. We utilize the LSTM network~\cite{hochreiter1997long} to track the temporal relationship across $T$ global vectors and then send the updated hidden state into a final MLP to regress the classification scores $S^a=\{s_i^a\}_{i=1}^K$ for $K$ classes. The HAR network can be supervised with a simple cross-entropy loss that measures the difference between $S^a$ and a one-hot ground truth activity label. 

\noindent\textbf{HP Base Network.} Similar to HAR, HP network also takes sequential radar point clouds as input and utilizes local encoders as well as global aggregation to obtain the latent representation and a global feature vector for each frame. Differently, the human parsing task needs to estimate per-point classification scores for every point rather than the whole sequence. Therefore, after applying the temporal information propagation, we use the point-level final features to regress the parsing scores $S^p=\{s_i^p\}_{i=1}^V$ for each point. Here $V$ denotes the number of human body segments. Our HP network is also supervised using a cross-entropy loss with per-point parsing labels. Specifically, we average the loss values over all classes to balance their impact on training.
\begin{table}[!t]
\centering
\begin{minipage}[t]{0.45\linewidth} 
    \centering
    \renewcommand\arraystretch{1.0}
    \setlength\tabcolsep{4pt}
    \caption{\small{Comparison of global aggregation mechanisms.}}
    \resizebox{\columnwidth}{!}{%
    \begin{tabular}{@{}lcccccc@{}}
        \toprule
        & \multicolumn{3}{c}{EPE3D (m) $\downarrow$} & \multicolumn{2}{c}{Acc3D $\uparrow$}\\
        \cmidrule(r){2-4} \cmidrule(r){5-6}
        Mechanism & ~All~ & ~Moving~ & ~Static~ & Strict & Relax    \\
        \midrule 
        Attention-based &  \textbf{0.046} &	\textbf{0.051} &	\textbf{0.009} &	 \textbf{0.406} & \textbf{0.703}\\
        Max-pooling & 0.050	& 0.059	& 0.011	 & 0.394 & 0.679\\
        Average-pooling & 0.047	& 0.055	& 0.009	 & 0.399 & 0.698\\
        \bottomrule
    \end{tabular}
    }
    \label{tab:mechanism}
    \vspace{-1em}
\end{minipage}
\hspace{0.5cm} 
\begin{minipage}[t]{0.45\linewidth} 
    \caption{\small{Results for different test environments.}}
    \renewcommand\arraystretch{1.0}
    \setlength\tabcolsep{6pt}
    \resizebox{\columnwidth}{!}{%
    \begin{tabular}{@{}lcccccc@{}}
        \toprule
        & \multicolumn{3}{c}{EPE3D (m) $\downarrow$} & \multicolumn{2}{c}{Acc3D $\uparrow$}\\
        \cmidrule(r){2-4} \cmidrule(r){5-6}
        Scene & ~All~ & ~Moving~ & ~Static~ & Strict & Relax    \\
        \midrule 
          Hallway & 0.040 & 0.046 & 0.007	& 0.464	& 0.746\\
          Square & 0.042 & 0.049 & 0.009 & 0.435 & 0.722\\
          Parking lot & 0.051	& 0.058	& 0.011	& 0.374	& 0.671\\
        \bottomrule
    \end{tabular}
    }
    \label{tab:environment}
    \vspace{-1em}
\end{minipage}
\end{table}

\subsection{Scene Flow Application Strategies}
As told in the main text, we propose two strategies to harness the scene flow network as a plug-and-play module to the downstream HAR and HP network. We believe both can effectively improve downstream performance. 

\noindent\textbf{S1 - Point cloud decoration.} The first way is to directly take the estimated scene flow as point-level raw features and decorate each radar point with them. Such additional scene flow features can explicitly provide the full motion information of radar point clouds. We adopt a two-stage training manner to implement this strategy, which first end-to-end trains the scene flow network and then freezes it to provide raw scene flow features to the downstream network training. 

\noindent\textbf{S2 - Latent feature recycling.} Another way is to leverage the latent representations encoded by scene flow networks to enhance the low-quality radar point clouds. By guiding the network to estimate accurate scene flow, the learned representation can provide high-level spatial-temporal information to downstream networks in a specific aspect. We follow a joint learning fashion to implement this strategy that takes the final features from the scene flow network as the input features to the downstream network and trains these two networks \emph{jointly} with a combination of the scene flow and downstream loss functions. 

\section{Sensitivity Analysis}\label{analysis}

Here we investigate the sensitivity of our scene flow network against a) global aggregation mechanisms, b) test environments and c) the number of training subjects, and conduct separate analyses for them in the following.

\subsection{Impact of Global Aggregation Mechanism} Besides the attention-based mechanism, another two typical operations, i.e., max-pooling and average-pooling, are tested into our network for global feature aggregation. The comparison results are shown in Table~\ref{tab:mechanism}. From the table, the attention-based mechanism yields the best performance as it can learn to dynamically adjust the weights according to per-point local features. The average-pooling performs better than the max-pooling. We credit this to the ability of the average-pooling to retain the mean feature across all points, which is more robust to outliers. 

\subsection{Impact of Test Environment} In Table~\ref{tab:environment}, we report the results of our trained model on each scene individually. As we can see, our model shows the best performance for the hallway scene while performing worst in the parking lot scene. We credit the difference in performance between test environments to two factors. First, the background clutter in the parking lot, is more complicated, which results in severe multi-path reflection during measurement. This decreases the SNR in radar data and further degrades the fidelity of generated radar point clouds. 
Second, our pseudo label generation is affected by the changeable outdoor illumination conditions. Not only does the 2D keypoint estimation on RGB images become more challenging, but also the strong sunlight can interfere with the depth measurements. Consequently, the quality of skeleton estimation in outdoor scenes is lower and the training is not as effective as in the indoor scene.

\begin{figure}[tbp!]
    \centering
    \includegraphics[scale=0.35]{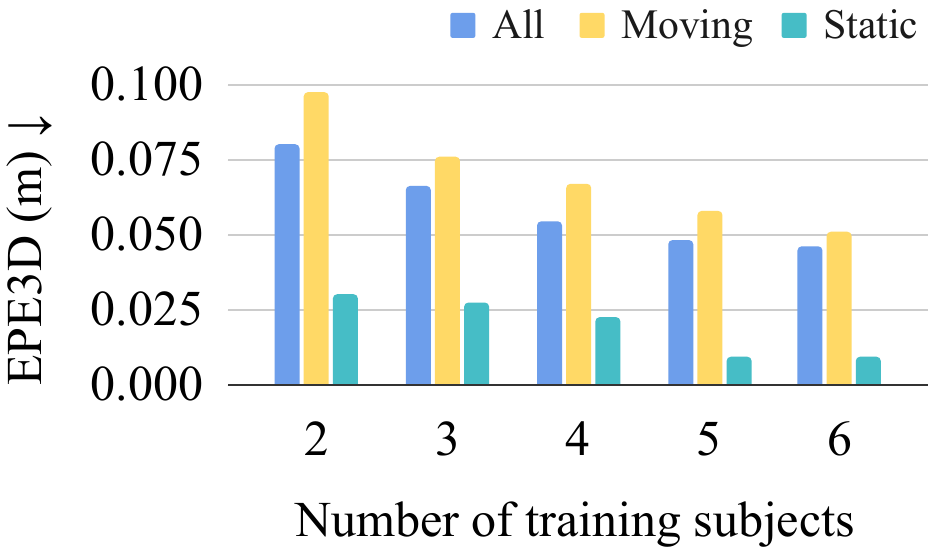}
    \hspace{0.5em}
    \includegraphics[scale=0.35]{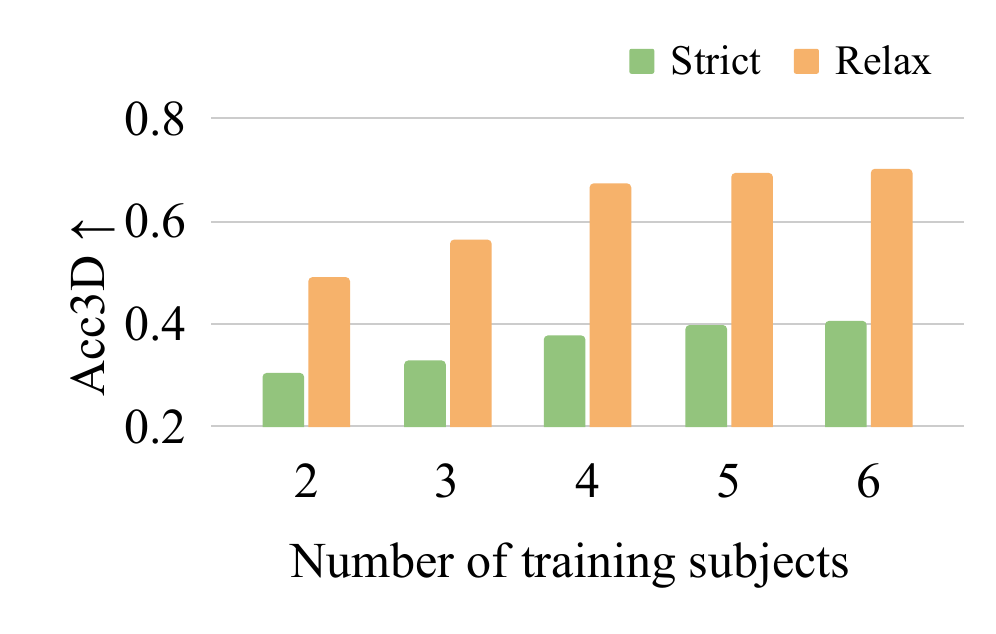}
    \caption{\small{Impact of the number of training subjects.}}
    \label{fig:num_subject}
    \vspace{-1em}
\end{figure}

\subsection{Impact of Training Subjects Number} In our experiments, the data from 6 subjects are used for training. Here we utilize 2/3/4/5 people out of the training set respectively for training to analyze the impact of the number of training subjects. As seen in Figure~\ref{fig:num_subject}, the performance of our network increases in all metrics as more subjects are added for training. This proves that there is still room to improve our performance when more training data is available. We can also observe that, even with only 3 training subjects (9k data frames), our network can still generalize well to the testing set with an overall EPE3D of 0.066m and a relax Acc3D of 0.56. This result further demonstrates the feasibility of our automatic scene flow labelling scheme to generate reliable pseudo labels.

\section{Limitation and Future Work}\label{limit}
Despite the progress achieved, there are limitations to be considered in future work. First, we only experiment on a stationary radar platform where only the human subject is movable. This is akin to the scenarios of human motion sensing in smart spaces and sensors are receded into the background environment. In future work, we plan to evaluate our system on mobile platforms, such as drones and wheeled robots, where the scene flow can be induced by both ego-sensor and subject motion. Second, as a proof-of-concept work, the functioning of our method is restricted to the single subject with face-forward body motion at this stage. We plan to extend our approach to multi-person scenarios and introduce variations in the subjects' position and orientation relative to the sensor in the next study. Lastly, the number of activities included in our dataset is limited as we only design activities that can be separated into the rigid motion of multiple body skeletons. This fits our need to annotate scene flow labels based on body skeleton displacement. In our upcoming work, we will design more complex and smaller activities and exploit a novel approach to derive the scene flow.

\bibliographystyle{splncs04}
\bibliography{main}
\end{document}